\documentclass[conference]{IEEEtran}
\usepackage{times}

\IEEEoverridecommandlockouts 

\usepackage{url}  %
\usepackage{graphicx}  %
\usepackage{subcaption}
\usepackage{xspace} %
\usepackage{multicol} %
\usepackage{multirow} %
\usepackage{hyperref} %
\hypersetup{bookmarksopen,bookmarksnumbered,
pdfpagemode=UseOutlines,
colorlinks=true,
urlcolor=black,
linkcolor=red,
anchorcolor=red,
citecolor=red,
filecolor=red,
menucolor=red
}
\usepackage{booktabs}
\usepackage{amssymb, amsfonts, amsthm}
\usepackage{commath}
\usepackage{nicefrac}
\usepackage{dsfont}
\usepackage{microtype}
\usepackage{mathtools}
\usepackage{xfrac}
\usepackage{verbatim}
\usepackage{xcolor}
\usepackage{tikz}
\usepackage{algorithm}
\usepackage[noend]{algpseudocode}
\usepackage{cleveref}
\usepackage{enumitem}
\usepackage[export]{adjustbox}
\usepackage{color}
\usepackage[textsize=tiny]{todonotes}
\usepackage[binary-units]{siunitx}
\usepackage{doi}

\pdfoutput=1

\newcommand{\xxnote}[3]{}
\ifx\hidenotes\undefined
 \usepackage{color}
 \renewcommand{\xxnote}[3]{\color{#2}{#1: #3}}
\fi

\usepackage[bibstyle=ieee,citestyle=numeric-comp,doi=false,natbib=true,maxcitenames=1,mincitenames=1]{biblatex}

\usepackage{flushend}

\newtheorem{proposition}{Proposition}

\newcommand{\agents}[0]{N}
\newcommand{\tasks}[0]{K}
\newcommand{\horizon}[0]{T}
\newcommand{\window}[0]{W}
\newcommand{\agent}[0]{n}
\newcommand{\task}[0]{k}

\definecolor{mygrey2}{RGB}{217,217,217}
\definecolor{mygrey3}{RGB}{189,189,189}
\definecolor{mygrey4}{RGB}{150,150,150}
\definecolor{mygrey5}{RGB}{99,99,99}
\definecolor{myorange}{RGB}{253, 174, 107}

\begin{document}

\title{
Dynamic Multi-Robot Task Allocation under Uncertainty and Temporal Constraints
}

\author{Shushman Choudhury, Jayesh K. Gupta, Mykel J. Kochenderfer, Dorsa Sadigh, and Jeannette Bohg
\thanks{All authors are with Stanford University, CA, USA. Please send correspondence to \texttt{shushman@cs.stanford.edu}}
}

\maketitle

\begin{abstract}
    We consider the problem of dynamically allocating tasks to multiple agents under time window constraints and task completion uncertainty. Our objective is to minimize the number
    of unsuccessful tasks at the end of the operation horizon.
    We present a multi-robot allocation algorithm that decouples the key computational challenges of sequential decision-making under uncertainty and multi-agent coordination and addresses them in a hierarchical manner.
    The lower layer computes policies for individual agents using dynamic programming with tree search, and
    the upper layer resolves conflicts in individual plans to obtain a valid multi-agent allocation.
    Our algorithm, {\em Stochastic Conflict-Based Allocation\/} (SCoBA), is optimal in expectation
    and complete under some reasonable assumptions. In practice, SCoBA is computationally efficient enough to interleave planning and execution online. On the metric of successful task completion, SCoBA consistently outperforms a number of baseline methods and shows strong competitive performance against an oracle with complete lookahead. It also scales well with the number of tasks and agents. We validate our results over a wide range of simulations on two distinct domains: multi-arm conveyor belt pick-and-place and multi-drone delivery dispatch in a city.
\end{abstract}

\section{Introduction}
\label{sec:intro}

\begin{figure}[t]
    \centering
    \begin{subfigure}{0.45\textwidth}
        \centering
        \includegraphics[width=0.95\textwidth]{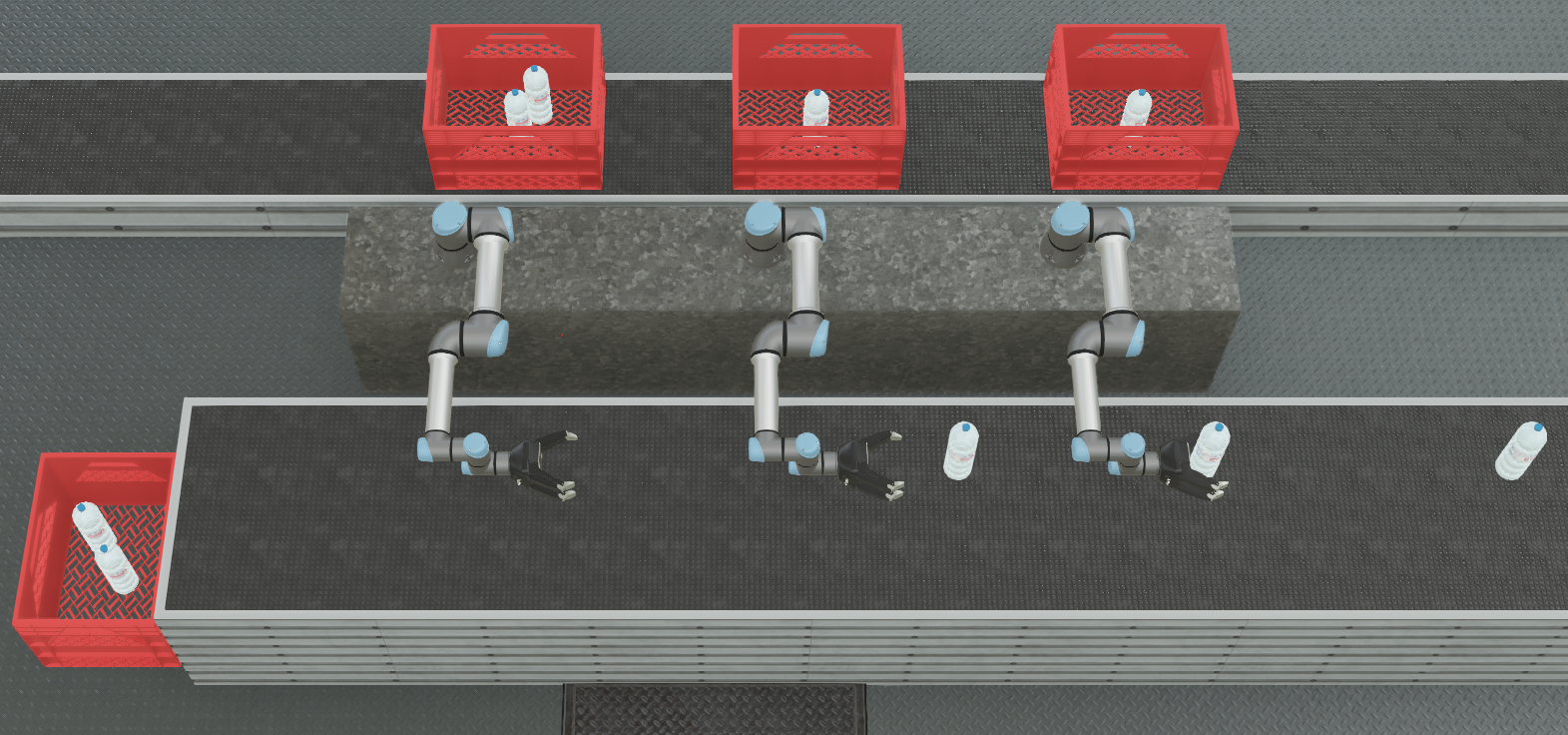}
        \caption{Conveyor Belt Pick-and-Place}
        \label{fig:page1_fig_conv}
    \end{subfigure}
    \begin{subfigure}{0.45\textwidth}
        \centering
        \includegraphics[trim=0 25 0 25, clip,width=0.95\textwidth]{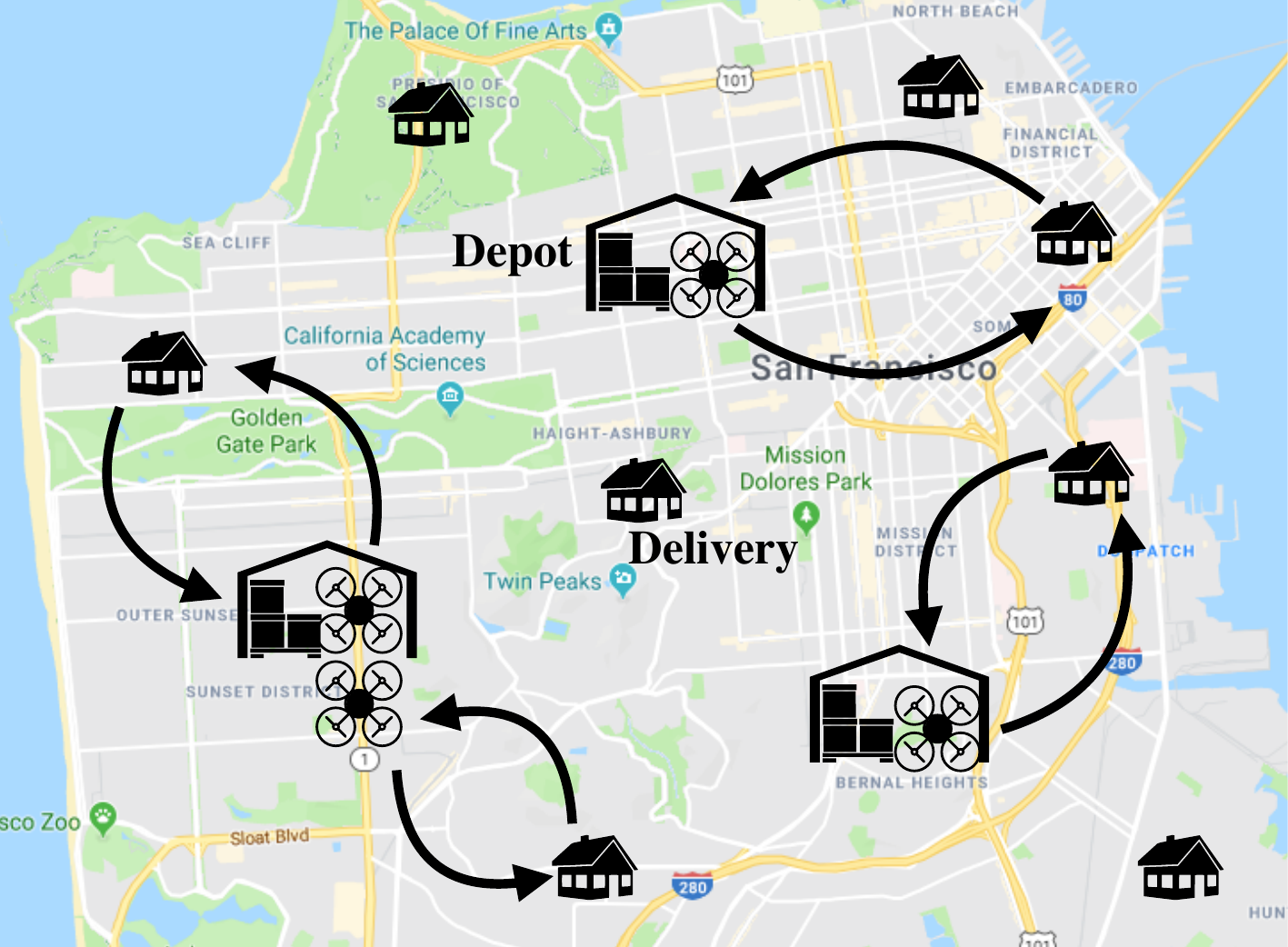}
        \caption{Multi-Drone Delivery Dispatch}
        \label{fig:page1_fig_drone}
    \end{subfigure}
    \caption{The above domains motivate our multi-robot task allocation approach.
    We allocate robots (arms or drones) to tasks (pick-and-place or delivery) that arrive online. Task completion is subject to uncertainty (grasping or flight time) and time window constraints.}
    \label{fig:page1}
    \vspace{-10pt}
\end{figure}

Efficient and high-quality task allocation is crucial for modern
cooperative multi-robot applications~\cite{DBLP:journals/ijrr/GerkeyM04}.
For warehouse logistics, teams of mobile robots carry goods between assigned locations~\cite{DBLP:conf/is/YanJC12}.
Industrial and manufacturing operations involve manipulators collaborating on assembly lines~\cite{johannsmeier2016hierarchical}.
On-demand ridesharing and delivery services
dispatch agents to incoming requests~\cite{hyland2018dynamic}.
Multi-robot task allocation needs to be computationally efficient and produce high-quality solutions under the challenges of real-world robotics: uncertainty of task execution success, temporal constraints
such as ordering and time windows, and tasks dynamically appearing online.
For instance, in one of our simulation domains, a team of robot arms pick
objects that appear on a conveyor belt from an external loading process and place them in bins (\Cref{fig:page1_fig_conv}).
With time window constraints induced by workspace limits, and uncertainty due to
imperfect grasping, the arms attempt to pick-and-place as many objects as possible.

Multi-robot task allocation is a difficult problem; it inherits the combinatorial optimization challenges of classical allocation
as well as the uncertainty and dynamic nature of robotics. 
Time-extended tasks and time window constraints 
further require algorithms to plan over horizons rather than
instantaneously, and account for spatio-temporal relationships among tasks~\cite{DBLP:conf/aaai/Gini17}.
The robotics community has worked on multi-agent task allocation with Markov Decision Processes~\cite{campbell2013multiagent} and robust task matching~\cite{liu2011assessing}. The classic multi-robot task allocation problem
has been classified extensively~\cite{DBLP:journals/ijrr/GerkeyM04} and
extended to account for uncertainty~\cite{DBLP:journals/arobots/MataricSO03}, temporal and ordering constraints~\cite{DBLP:conf/aaai/Gini17},
and dynamic task arrivals~\cite{DBLP:journals/anor/CordeauL07} (some of these will be baselines for our method). The operations research community has developed methods for task execution uncertainty~\cite{timotheou2010asset,rahmani2014robust} and online rescheduling~\cite{o1999predictable}.
However, their simplified agent models (such as flow shops) do not address the
spatial relationships between tasks or the intersection of uncertainty and time constraints.

The algorithmic challenges for our allocation setting are \emph{sequential planning under uncertainty} and \emph{coordinating multi-agent decisions}. Prior works typically attempt the computationally intractable joint problem and thus require simplifying approximations or heuristics for either planning or coordination.
Our key idea is to \emph{decouple these challenges and address them hierarchically} in an efficient two-layer approach. The lower layer plans for individual agents, using dynamic programming on a policy tree to reason about the uncertainty over task completion. The upper layer uses optimal conflict resolution logic from the path planning community to coordinate the multi-agent allocation decisions~\cite{sharon2012conflict}. Our overall algorithm, {\em Stochastic Conflict-Based Allocation} (SCoBA), yields allocation policies that minimize the expected cumulative penalty for unsuccessful tasks; SCoBA can also seamlessly interleave planning and execution online for new tasks.

The following are the contributions of our work:
\begin{itemize}
    \item We propose a general formulation for multi-robot allocation under task uncertainty and temporal constraints.
    \item We present a hierarchical algorithm, SCoBA, which uses multi-agent conflict resolution with single-agent policy tree search. We prove that SCoBA is both optimal in expectation and complete under mild assumptions.
    \item 
    We demonstrate SCoBA's strong competitive performance against an oracle with complete lookahead, and its performance advantage over four baseline methods for successful task execution. Our results also show that SCoBA scales with increasing numbers of agents and tasks. We run simulations on two distinct robotics domains (\Cref{fig:page1}); a team of robot arms picking and placing objects from a conveyor belt and on-demand multi-drone delivery of packages in a city-scale area.
\end{itemize}
\section{Background and Related Work}
\label{sec:related}

We briefly discuss three background areas: i) algorithms for assignment and scheduling, ii) multi-agent
decision-making, and iii) relevant work in multi-robot task allocation.
\\
\noindent
\textbf{Assignment and Scheduling:}
A major topic in discrete optimization is assigning resources to tasks~\cite{DBLP:books/daglib/0022248}, for which the Hungarian algorithm is a
fundamental approach~\cite{munkres1957algorithms}.
In temporal tasks, we use the \emph{scheduling} model,
where the objective is a function of completed jobs~\cite{pinedo2012scheduling}.
Scheduling problems with multiple resources and tasks are computationally hard, even when deterministic~\cite{lenstra1977complexity}. In online scheduling, each task is only observed when made available~\cite{albers1999better}. Hard real-time tasks have a time window constraint for completion~\cite{dertouzos1989multiprocessor}.
Approaches for scheduling under uncertainty address problems where task execution is not fully deterministic~\cite{DBLP:conf/icadlt/ChaariCAT14,sadigh2016safe}. These approaches are either proactive in anticipating future disruptions~\cite{lin2004new}, reactive to changes~\cite{szelke1994knowledge,raman2015reactive}, or hybrid~\cite{church1992analysis}. For scenarios with ordering constraints (such as assembly lines), additional models like job shops~\cite{al2011robust} and flow shops~\cite{gonzalez2017flow} are useful, particularly real-time flow shop scheduling~\cite{rahmani2014robust,framinan2019using}.
Iterative conflict resolution, which we adapt in our approach, was used for scheduling~\cite{gay2015conflict}, joint task assignment and pathfinding~\cite{DBLP:conf/atal/MaK16}, and diagnosis and repair~\cite{ghosh2016diagnosis}.
This extensive body of work provides valuable insights but does not consider spatial relationships between tasks and their coupling with temporal uncertainty.
\\
\noindent
\textbf{Multi-Agent Sequential Decision-Making:}
The Markov Decision Process (MDP) is a mathematical model for our setting of sequential decision making under uncertainty~\cite{kochenderfer2015decision}.
Different solution techniques exist for MDPs, depending on available information; dynamic programming~\cite{bertsekas2005dynamic} when the explicit transition model is known, sample-based online planning~\cite{peret2013online} when only a generative model exists, and reinforcement learning~\cite{sutton2018reinforcement} when no model is available. Our problem is a multi-agent MDP (MMDP), where agents coordinate to achieve a single shared objective; planning for MMDPs is generally computationally intractable due to the exponentially large decision space~\cite{boutilier1996planning}. Although reinforcement learning techniques are often employed to alleviate some tractability issues~\cite{littman1994markov,lanctot2017unified} by learning values of different states and actions, model-free methods like Q-Learning face exploration challenges~\cite{lanctot2017unified}. Online tree search methods can fare better by focusing on relevant states more effectively~\cite{Vodopivec2017-oh}. As we will demonstrate, merely framing multi-robot task allocation as an MMDP will not yield good quality solution strategies (due to its computational challenges).
\\
\noindent
\textbf{Multi-Robot Task Allocation:}
We outline a number of domain-agnostic and domain-specific works on multi-robot task allocation.
MDP solvers have been used to generate a sequential greedy strategy, but without accounting for completion uncertainty~\cite{campbell2013multiagent}. The probability of task failure has been considered by two-stage Stochastic Integer Programs and network flow algorithms, which are exhaustive combinatorial approaches unsuitable for tasks streaming in online~\cite{ahmed2003dynamic,timotheou2010asset,timotheou2011network}. A sensitivity analysis approach to optimal assignment under uncertainty provides some insights on robustness but has no notion of temporal constraints~\cite{liu2011assessing}.
The taxonomies for multi-robot task allocation under temporal constraints help us characterize our problem's difficulty~\cite{DBLP:journals/ijrr/GerkeyM04,DBLP:journals/ras/NunesMMG17}. 

Previous work on collaborative assembly lines includes hierarchical planning
frameworks, constraint programming, and robust scheduling for robotic flowshops~\cite{johannsmeier2016hierarchical,behrens2019constraint,che2017efficient}.
However, they all simplify one or more key complexities such as task completion uncertainty
or multi-agent configuration models.
Dynamic vehicle dispatch problems have been explored in work on vehicle routing algorithms with time
windows and trip assignment algorithms for ridesharing~\cite{DBLP:journals/eor/LauST03,alonso2017demand}.
However, they make restrictive assumptions on the uncertainty and environment dynamics~\cite{DBLP:conf/aaai/Gini17}.
Driver-task assignment with uncertain durations and task windows do solve for a similar setting as ours but assume some knowledge of future requests~\cite{cheung2005labeling}. 

\section{Problem Formulation}
\label{sec:problem}

We base our formulation on previous work for multi-robot task allocation with temporal constraints~\cite{DBLP:conf/aaai/Gini17}.
There is a set of $\agents$ \textbf{agents}, denoted as $[\agents]$ and $\tasks$ \textbf{tasks}, denoted as $[\tasks]$; the problem horizon is $\horizon$ time-steps. For each agent $\agent \in [\agents]$ and task
$\task \in [\tasks]$, the service \textbf{time window} is $\window_{\agent\task} = \left( t_{\agent\task}^l, t_{\agent\task}^u\right)$, where $l$ and $u$ are respectively the lower and upper time limits within which $\agent$ can
attempt $\task$. 
There may also be an additional so-called \textbf{downtime} if the agent executes the task successfully, e.g., the time for a robot arm to transfer the object to the bin.
We represent \textbf{task duration uncertainty} as
\begin{equation}
    \label{eq:cumul}
    \tau_{\agent\task}(t) = \mathrm{Prob}\left[\agent \text{ completes } \task \text{ within } t \text{ time-steps } \right].
\end{equation}
We assume knowledge of this cumulative distribution as part of the problem specification, typical for task scheduling under uncertainty~\cite{DBLP:conf/icadlt/ChaariCAT14}; the particular model is domain-dependent.
By definition, the conjunction of $\window$ and $\tau$ imposes an upper bound on \textbf{task completion probability}, i.e.,
\begin{equation}
    \label{eq:pdf}
    \mathrm{Prob}\left[\agent \text{ completes } \task\right] \leq \tau_{\agent\task}(t_{\agent\task}^u - t_{\agent\task}^l).
\end{equation}
For all unsuccessful tasks, the system incurs a \textbf{penalty} of $\sum_k J(\task)$ units.
An agent can attempt only one task at a time.

We seek an allocation policy that \emph{minimizes the expected cumulative penalty due to unsuccessful tasks}. An allocation policy $\pi$ is a mapping from the agents to the tasks and
their respective attempt times, i.e. $\pi : [\agents] \rightarrow [\tasks] \times [\horizon]$.
Since there is uncertainty about task completion, a single-shot allocation is insufficient.
Of course, the attempt times for future tasks depend on when the earlier tasks are actually executed (successfully or unsuccessfully). Our optimization problem is
\begin{equation}
    \label{eq:opt-prob}
    \begin{aligned}
        & \underset{\pi \in \Pi}{\mathrm{argmin}}
        & & \mathbb{E} \Big[\sum_{\task \in [\tasks]} \mathds{1}[k] \cdot J(\task) \mid \pi \Big]\\
        & \text{s.t.}
        & & t \in \window_{\agent\task} \ \ \forall \ (\task,t) \in \pi(\agent),
    \end{aligned}
\end{equation}
where the indicator function $\mathds{1}[k] = 1$ if the task $k$ remains incomplete at the end of the horizon, and $\Pi$ is the set of all possible allocation policies.
The constraint enforces that an agent attempts a task within the valid time window.
The expectation is over the task execution success distribution for the allocation policy.
For the rest of the discussion, we will assume that $J(\task) = 1$, i.e., all tasks are equally important; this objective is the \emph{unweighted tardy jobs penalty}~\cite{pinedo2012scheduling}.

\begin{figure}
    \centering
    \includegraphics[width=0.95\columnwidth]{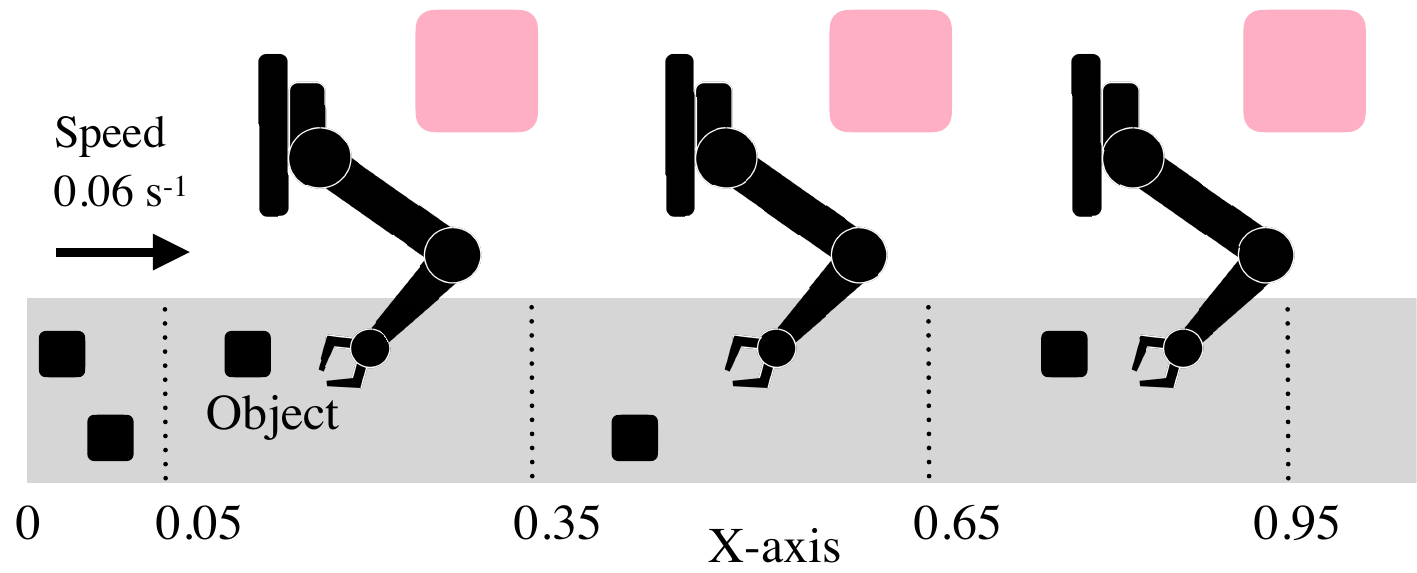}
    \caption{The illustrated conveyor belt has $\agents = 3$ arms and $K = 5$ objects. The belt is of unit length, and each arm's workspace spans $0.3$ units (dashed lines are the limits). Given the belt speed, the agent-task time window for any arm-object pair is at most $\SI{5}{\second}$.}
    \label{fig:beltviz}
    \vspace{-10pt}
\end{figure}

The discrete-time rolling horizon formulation described above is fairly general and useful.
We are concerned with high-level allocation rather than the underlying task execution, so we avoid the added complexity
of continuous-time representations. The underlying tasks typically involve time-constrained trajectory planning, for which there are well-established models and methods~\cite{laumond1998robot}.
Furthermore, we can interleave planning and execution suitably and recompute an allocation policy when new tasks appear online (and we do so in practice).
\\
\noindent
\textbf{Motivating Examples:}
We describe two distinct robotics settings to instantiate our formulation.
First, consider the previously introduced example of robot arms along a conveyor belt (see~\Cref{fig:beltviz}).
Each arm has an associated collection bin for objects picked up from the belt. The objects appear on the belt through an external process. The arms take varying amounts of time for picking, depending on the quality of the grasp strategy or gripper attributes. Arms have finite reach, and an object may not be picked up before it goes out of reach. Objects missed by all arms must be sorted by hand afterwards. The goal is to successfully pick-and-place as many objects, or equivalently, miss as few objects as possible.

Second, consider on-demand multi-drone dispatch for package delivery in a city (note the underlying similarities to the previous example). Delivery tasks arise through an external process of customer requests. Drones take varying amounts of time to travel from the product depot to the package delivery location, depending on flight conditions.
Requests arrive with time windows, such that drones must wait until the window starts
to deliver the product to the customer, and late deliveries are penalized. Over a certain time horizon,
our objective is to minimize the number of late deliveries.
\\
\noindent
\textbf{Challenges:} To motivate our approach, we briefly discuss the problem complexity.
By the multi-robot task allocation taxonomy of~\citet{DBLP:journals/ijrr/GerkeyM04}, the deterministic version of our problem is ST-SR-TA, i.e. a single robot (SR) executes a single task (ST) at a time, where tasks
are time-extended (TA) rather than instantaneous. ST-SR-TA problems are an instance of an NP-Hard scheduling problem, specifically multi-agent scheduling with resource constraints~\cite{garey1975complexity}.
The uncertainty of task execution success exacerbates this difficulty.
Time windows make allocation harder by requiring algorithms to account for spatio-temporal task relationships~\cite{DBLP:conf/aaai/Gini17}. Finally, new tasks streaming in require our approach to interleave planning and execution effectively, e.g., by replanning at task arrivals~\cite{DBLP:journals/anor/CordeauL07}.
\begin{figure*}[t]
    \centering
    \includegraphics[width=0.9\textwidth]{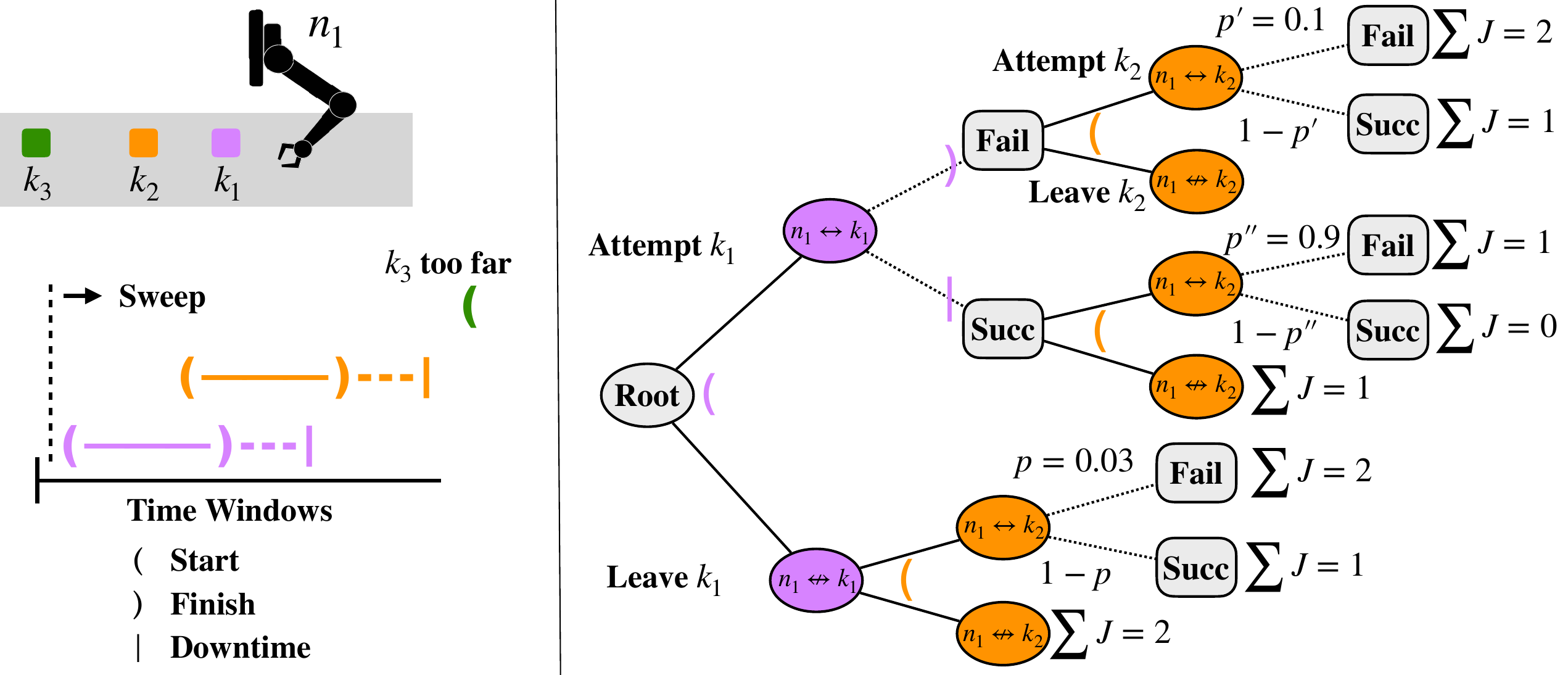}
    \caption{The low-level routine of SCoBA generates the policy tree over valid tasks for an individual agent, specifically, by sweeping along the time axis and branching on the start or finish of a task's time window. At the start of a window, two new \emph{decision nodes} (ovals) are introduced: \emph{to attempt} ($\leftrightarrow$) or \emph{to leave} ($\not\leftrightarrow$) the task respectively. At the end of a time window and the downtime, the \emph{outcome nodes} (rectangles) depict failure or success. After the tree generation, dynamic programming propagates the values from the leaves to the root. The probability values $p = 0.03$, $p' = 0.1$, $p'' = 0.9$ are just hypothetical values that illustrate how the same attempt node $(\agent_1 \leftrightarrow \task_2)$ has three different copies, with different outcome probabilities (depending on the branch of the tree).}
    \label{fig:tree}
    \vspace{-10pt}
\end{figure*}

\section{Hierarchical Multi-Robot Task Allocation}
\label{sec:approach}

Our key algorithmic challenges are \emph{sequential planning under uncertainty} (of task completion) and \emph{multi-agent coordination} (of allocations). The joint multi-agent planning problem is  computationally prohibitive for large settings~\cite{boutilier1996planning}; most closely related previous works either use simplifying approximations for planning and optimization~\cite{timotheou2011network,hyland2018dynamic} or simple coordination heuristics~\cite{DBLP:conf/aaai/KartalNGG16,DBLP:journals/arobots/MataricSO03}.
In contrast, \emph{we address the challenges hierarchically in a two-layer approach} called
Stochastic Conflict-Based Allocation (SCoBA).
At the low level, we independently determine the optimal task attempt sequence for each individual agent, ignoring other agents. At the high level, we resolve potential conflicts in assigned tasks across multiple agents to obtain a valid
multi-robot allocation. In this section, we will discuss in detail the two layers and how they come together in SCoBA. We will then briefly mention
how we interleave planning and execution online and how SCoBA can exploit sparse agent interactions using coordination graphs.

\subsection{Low-Level: Single Agent Policy}
\label{sec:approach-tree}

We consider the perspective of an individual agent, independent of the other ones.
From the definition in~\Cref{sec:problem}, 
we have the set of current tasks, corresponding time windows, and task completion uncertainty distribution, and we want a task attempt policy tree for the agent.
Since task execution is stochastic, the first possible attempt time of a task depends on the tasks attempted before it. We make a simplifying approximation -- \emph{the agent attempts a task as soon as possible and observes the outcome at the end of the window}. This approximation collapses the temporal dimension by treating tasks as discrete events
rather than extended ones.

We illustrate the policy tree search process for a single robot $\agent_1$ and three tasks (objects) $\task_1, \task_2, \task_3$ in~\Cref{fig:tree}.
First, we sort tasks in increasing order of the start of their time window. Then, we sweep along the
time axis and update the tree at every \emph{event point}, i.e., the start or finish of the window (and the end of the downtime if the task were to be successful). 
The updates to the policy tree depends on the event point (start/finish/downtime). For the start of a time window,
we introduce two new \emph{decision nodes} (ovals) to attempt ($\leftrightarrow$) or leave ($\not\leftrightarrow$) the task respectively.
At the end of a time window and the downtime, we introduce \emph{outcome nodes} (rectangles)
respectively for failure or success, where the outcome probability $p$ depends on the minimum feasible start time for the attempt, which in turn depends on the specific branch of the tree. For instance, notice in~\Cref{fig:tree} the three copies of the decision node ($\agent_1 \leftrightarrow \task_2$), with different probabilities, depending on whether it was attempted after the failure, success or non-attempt of task $\task_1$.

The leaves of the binary policy tree contain the cumulative penalty along their branches, e.g., a penalty of $1$ for each unsuccessful task. 
We then use dynamic programming to propagate values upwards from the leaves to the root.
For a pair of outcome node siblings, we set the parent's value (denoted as $V$) to the expected value of its children,
\begin{equation}
    V(\mathrm{parent}) \ := \ p \cdot V(\mathrm{Fail}) + (1-p) \cdot V(\mathrm{Succ}).
\end{equation}

For a pair of decision node siblings, the parent's value is the minimum of the children's, i.e.,
\begin{equation}
    V(\mathrm{parent}) \ := \min \{V(\mathrm{child1}), \ V(\mathrm{child2})\};
\end{equation}
in the running example in~\Cref{fig:tree}, we have $V(\text{root}) = \min \{V(\agent_1 \leftrightarrow \task_1), V(\agent_1 \not\leftrightarrow \task_1)\}$.
The resulting tree encodes the policy that minimizes the agent's expected penalty for
all tasks up to the planning horizon, and $V(\text{root})$ is the value of this expected penalty. We obtain the next task assigned to the agent by following child nodes of minimum value until the first attempt node (e.g., $\agent_1 \leftrightarrow \task_1$).

\subsection{High-Level: Multi-Agent Coordination}
\label{sec:approach-coordination}

The policy tree determines the approximately optimal task attempt sequence
for an individual agent (approximate due to the temporal simplification mentioned earlier).
The tree searches are independent of each other, so two agents may have conflicting allocations.
Since our objective function depends on all agents, \emph{breaking ties na\"ively could yield arbitrarily poor global allocations}. 
Multi-agent pathfinding algorithms face a similar challenge and have
to resolve inter-agent conflicts between shortest paths~\cite{felner2017search}. Conflict-Based Search is an
effective strategy for this problem~\cite{sharon2012conflict}; by decoupling single-agent path planning and inter-path conflict resolution, it is efficient in practice without losing optimality.

\begin{figure}[t]
    \centering
    \includegraphics[width=0.9\columnwidth]{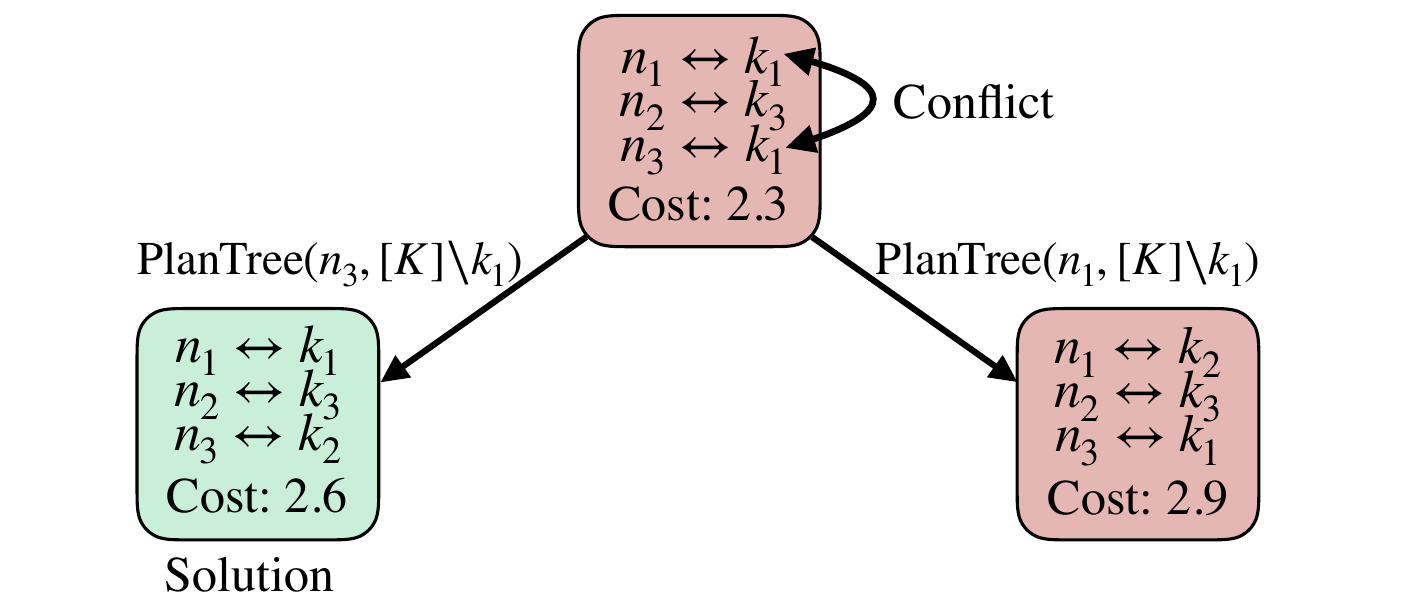}
    \caption{A constraint tree node with a conflict in the allocation generates two children with corresponding constraints
    on the conflicting agents ($\agent_1$ and $\agent_3)$ and task ($\task_1)$. Best-first search on the constraint tree returns the first high-level node with a conflict-free allocation.}
    \label{fig:cbs}
    \vspace{5pt}
\end{figure}

We leverage the idea of inter-agent conflict resolution from Conflict-Based Search. The high level of our algorithm, SCoBA,
searches a binary \emph{constraint tree} (\Cref{fig:cbs}) generated from conflicts between solutions for individual
agents obtained from the low level, i.e., the policy tree search.
Two agents $n_1$ and $n_2$ are in \emph{conflict} if they are allocated the same task $k$ in overlapping
time windows, i.e., if $(\task,t_1) \in \pi(\agent_1), (\task,t_2) \in \pi(\agent_2)$ and either
$t_2 \in \window_{\agent_1,\task}$ or $t_1 \in \window_{\agent_2,\task}$.
A \emph{constraint} for an agent is a task excluded from consideration by the tree search for that agent. 
Each node in the constraint tree maintains (i) a set of constraints, i.e., tasks to ignore, for each agent, (ii) a multi-agent allocation that respects all constraints, and (iii) the cost of the allocation. For SCoBA, the cost of the allocation is the sum of expected penalties for each agent, where the expected penalty for each agent is the value of the root node of its policy tree. The allocation cost is used as the criteria for \emph{best-first search on the constraint tree}; this best first search
continues until it finds a conflict-free allocation.

\subsection{Stochastic Conflict-Based Allocation (SCoBA)}
\label{sec:approach-scoba}

\Cref{alg:SCoBA} describes SCoBA, using the tree search of~\Cref{sec:approach-tree}
as the \textsc{PlanTree} subroutine. Its structure is similar to a
presentation of Conflict-Based Search by~\citet{felner2017search}.
The constraint tree is initialized with the root node, which has an empty constraint set and the allocation from running \textsc{PlanTree} for each individual agent (\crefrange{line:init}{line:insert}).
When a high-level node is expanded, the corresponding allocation is checked for validity (\cref{line:valid}). If there is no conflict, this allocation is returned as the solution. Otherwise, for every conflict between two or more agents, new child nodes are added, where constraints are
imposed on the agents involved (\cref{line:child}). 
A child constraint tree node inherits its parent's constraints and adds one more constraint for a particular agent.

Consider the simple illustrative example in~\Cref{fig:cbs}. The root node has agents $\agent_1$ and $\agent_3$ both assigned to task $\task_1$.
This conflict yields two constraints, one inherited by each of the two child nodes. The first constraint excludes $\task_1$ from the recomputed policy tree
search for $\agent_1$. The second constraint does the same for $\agent_3$. For each new (non-root) node, the low level tree search is only re-run on the agent for which the constraint is added (\cref{line:rerun}). Both of the resulting child nodes are conflict-free, but the left one, with a lower allocation
cost of $2.6$, is returned as the solution.
\\
\\
\noindent
Our problem setting is both online and stochastic. However, under some simplifying assumptions, we can establish optimality and completeness properties for SCoBA.

\setlength{\textfloatsep}{1pt}
\begin{algorithm}[t]
\caption{Stochastic Conflict-Based Allocation}
\begin{algorithmic}[1]
\Procedure{Main}{$[\agents], [\tasks], T, \window_{\agent,\task}, \tau_{\agent\task} \ \forall \ \agent, \task$}
    \State Initialize A as the root \label{line:init}
    \State $A.soln \gets \textsc{PlanTree}(\agent,[\tasks],T,\window,\tau)$ $\forall$ $n$ 
    \State $A.constr \gets \{ \}$ \Comment{Empty constraint set}
    \State $A.cost \gets \text{SumOfIndividualCosts}(A.solution)$
    \State Insert $A$ into \textsc{Open} \Comment{Open list of Constraint Tree} \label{line:insert}
    \While{\textsc{Open} not empty}
        \State $S \gets \mathrm{PopBest}(\textsc{Open})$ \Comment{Min. Cost Allocation}
        \If{$S.soln$ is valid}           \Comment{No conflicts} \label{line:valid}
            \State \textbf{return} $S.soln$
        \EndIf
        \State $C \gets \text{find-conflicts}(S)$     \Comment{Inter-agent conflicts}
        \For{all conflicts $(n,k) \in C$}
            \State $A \gets \textsc{GenerateChild}(S,n,k)$
            \State Insert $A$ into \textsc{Open} \label{line:child}
        \EndFor
    \EndWhile
\EndProcedure
\Statex
\Procedure{GenerateChild}{$S,n,k$}
    \State $A.constr \gets S.constr \cup k$  \Comment{Task to exclude}
    \State $A.soln \gets S.soln$
    \State $A.soln \gets \textsc{PlanTree}(n,[\tasks] \setminus A.constr,T,\window,\tau)$ \label{line:rerun}
    \State $A.cost \gets \text{SumOfIndividualCosts}(A.soln)$
    \State \textbf{return} $A$
\EndProcedure
\end{algorithmic}
\label{alg:SCoBA}
\end{algorithm}

\begin{proposition}
\label{prop:opt}
If (i) no new tasks are added online, (ii) the tree search is executed to the full horizon, and (iii) task completion is determined at the end of the time window, then SCoBA is optimal in expectation, i.e. SCoBA minimizes in expectation the number of incomplete tasks at the end of the time horizon.
\end{proposition}

\begin{proposition}
Under the assumptions of~\Cref{prop:opt}, SCoBA is complete. If a valid allocation exists, SCoBA returns it.
\end{proposition}

We provide detailed proofs in the appendix. We derive them
from the corresponding optimality and completeness proofs of the Conflict-Based Search algorithm for
multi-agent pathfinding~\cite{sharon2012conflict}.
For optimality, we use existing results in sequential decision-making
to show that SCoBA's low-level routine, i.e., policy tree generation, is optimal
in expectation for an individual agent~\cite{kochenderfer2015decision}.
We then prove that SCoBA's high-level multi-agent coordination
satisfies the sufficient condition to inherit the multi-agent optimality of Conflict-Based Search.
For completeness, we show how the high-level constraint tree of SCoBA, as in Conflict-Based Search, has a finite number of nodes. Therefore, systematic best-first search on it will find a valid solution if
one exists.
\\
\noindent
\textbf{Interleaving Planning and Execution:}
To account for new tasks beyond the horizon, we interleave planning and execution online.
SCoBA's elegant representation makes interleaving straightforward at both levels. For the single agent policy tree search,
we truncate the search horizon based on computation
requirements. In our implementation, we run the sweep until the first task whose time window begins after the downtime of all tasks before it ($\task_3$ in~\Cref{fig:tree}). For the multi-agent coordination,
we set a threshold on the number of high-level conflicts, once again based on real-time
computation constraints. If the threshold is exceeded, we return the
current high-level solution. For agents allocated to the same task, we break ties arbitrarily
and keep the unassigned agents for allocation to new tasks at the next timestep.
\\
\noindent
\textbf{Coordination Graphs:}
SCoBA works with any arbitrary configuration of agents and inter-agent constraints. We also use coordination graphs (CGs) from multi-agent decision-making for greater efficiency~\cite{kok2003multi}.
In CGs, each node represents an agent, and each edge encodes a (potentially directed) dependency between agents, such that only connected agents need to coordinate actions (or allocations in our case). 
The choice of coordination graph for a problem
is domain-dependent and often quite natural. For instance, in the conveyor belt
example, the arms are ordered along the belt and their workspaces are mutually
exclusive, therefore the coordination graph is a directed chain from the first arm to the last.

The CG structure impacts the high-level multi-agent coordination stage of SCoBA.
The \emph{absence of an edge between two agents implies that their sets of possible tasks are disjoint}, i.e.,
they cannot have conflicting allocations.
Therefore, in practice, SCoBA need not consider dependencies between all the agents.
If the CG is directed (as in the conveyor belt), we run the tree search for agents along a topological ordering of the CG. For any agent, we exclude the tasks already assigned to its predecessors. By construction, we will obtain a conflict-free allocation at the end (without any child nodes being generated in the high-level constraint tree).
If the CG is undirected (as in our multi-drone delivery domain), such a topological ordering is not feasible, and
conflicts may be unavoidable. However, if the CG has multiple connected components, then nodes (agents) in different components cannot conflict with each other, so we can run SCoBA on each component in parallel.

\section{Evaluation}
\label{sec:eval}

The primary metric for evaluating SCoBA is the accumulated penalty for unsuccessful tasks.
We will also evaluate its scalability to tasks and agents via computation time.
We first outline the range of methods we use to baseline SCoBA. We then
present and discuss the results for both performance metrics on simulations
for each of our two distinct robotics-inspired domains: conveyor belt pick-and-place
and on-demand multi-drone delivery dispatch.
We use Julia~\cite{Julia-2017} on a \SI{16}{\gibi\byte} RAM machine and a $6$-core \SI{3.7}{\giga\hertz} CPU for all simulations\footnote{The code is available at \url{https://github.com/sisl/SCoBA.jl}}. 
% The accompanying video illustrates scenarios for both domains.

\subsection{Baselines for Unsuccessful Task Penalty}
\label{sec:eval-baselines}

We use multiple complementary methods to baseline SCoBA on our primary metric of unsuccessful task penalty:
\begin{enumerate}
    \item \textbf{EDD}: The Earliest Due Date heuristic assigns each agent
    to the task with the nearest time window deadline and is a common heuristic for scheduling~\cite{pinedo2012scheduling}.
    \item \textbf{Hungarian}: An unbalanced Hungarian algorithm, where the edge weight for an agent-task pair is the probability of successful task completion~\cite{munkres1957algorithms}. This method is a special case of a general purpose network-flow approach, where only one task is assigned at a time~\cite{timotheou2011network}.
    \item \textbf{Q-Learning}: We frame the multi-robot task allocation problem as a discrete-time Markov Decision Process and pre-compute a policy with Q-Learning.
    \item \textbf{MCTS}: A recent Monte-Carlo Tree Search approach specifically for multi-robot task allocation~\cite{DBLP:conf/aaai/KartalNGG16}. The tree search is conceptually similar to ours (albeit with Monte Carlo sampling of outcomes) but it uses arbitrary priority orderings among agents to coordinate decisions and control the tree branching factor.
\end{enumerate}
For the baselines, we cover a range of approaches for multi-robot allocation from scheduling to sequential decision-making under uncertainty. Both EDD and Hungarian are reactive, i.e., do not plan sequentially. The latter optimizes for multiple agents, unlike the former. Both Q-Learning and MCTS plan sequentially by framing an MDP, but the former is model-free
and offline while the latter is model-based and online.

\subsection{Conveyor Belt: Experiments and Results}
\label{sec:eval-convbelt}

\begin{figure*}[t]
    \centering
    \begin{subfigure}{0.325\textwidth}
        \centering
        \includegraphics[width=\textwidth]{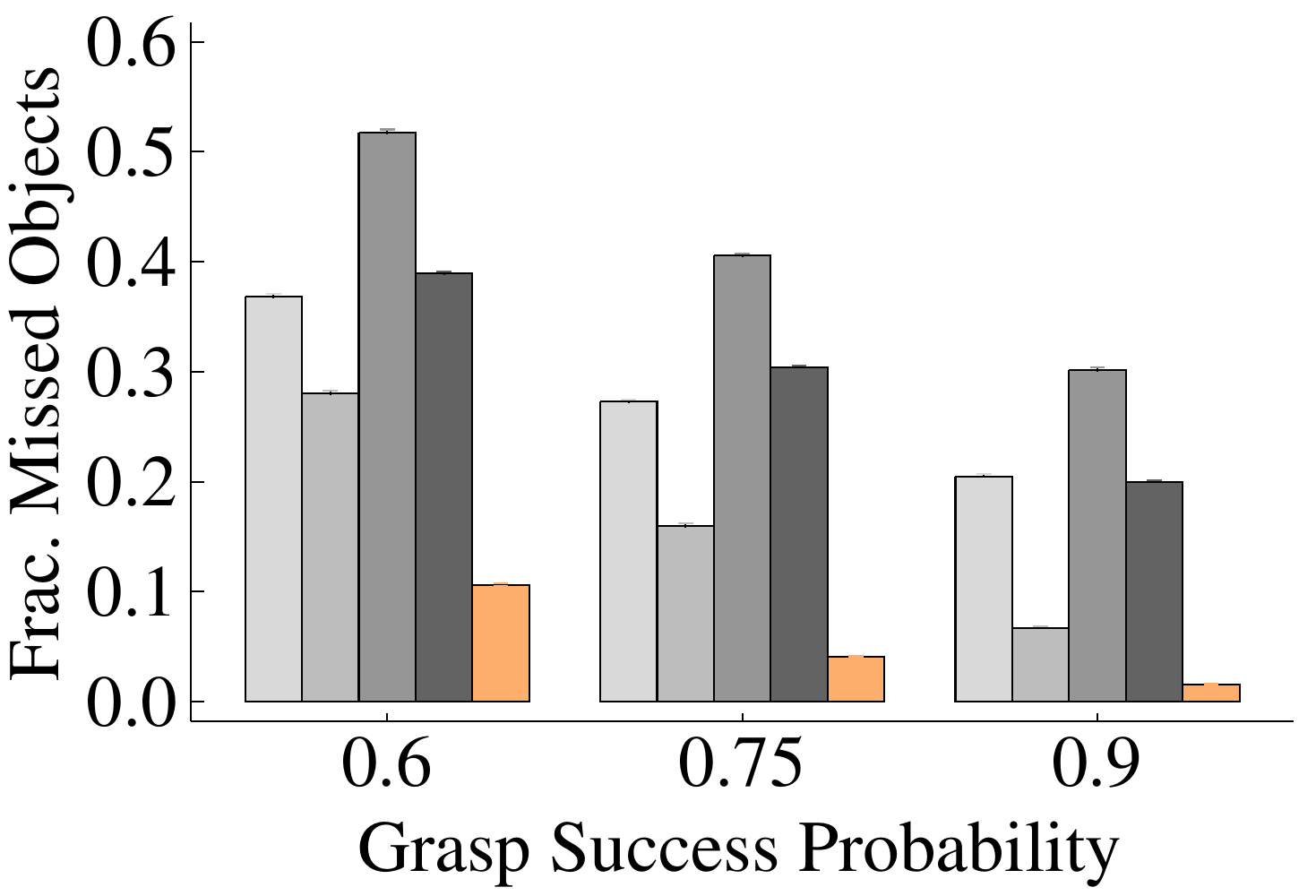}
        \label{fig:belt-perf-grasp}
    \end{subfigure}
    \begin{subfigure}{0.325\textwidth}
        \centering
        \includegraphics[width=\textwidth]{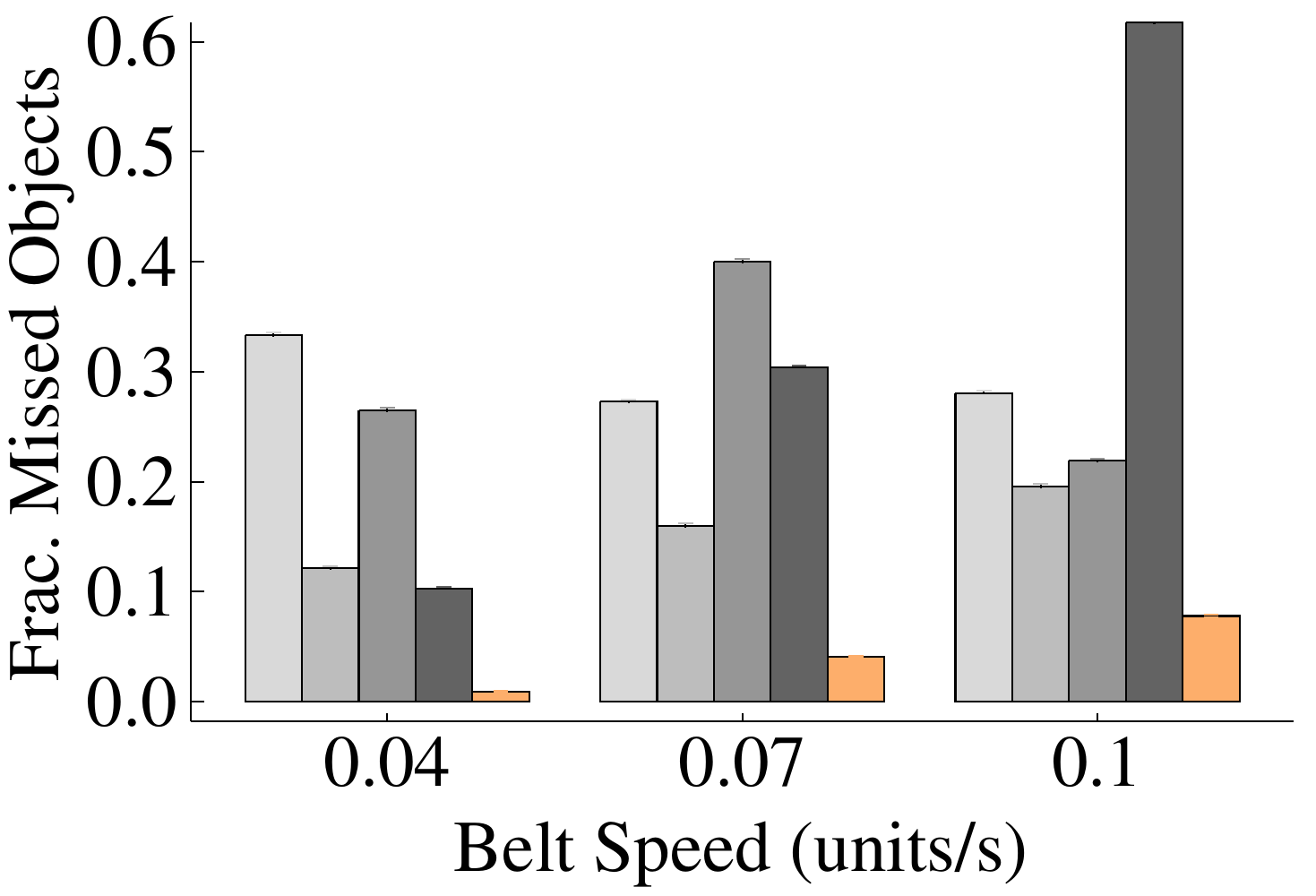}
        \label{fig:belt-perf-speed}
    \end{subfigure}
    \begin{subfigure}{0.325\textwidth}
        \centering
        \includegraphics[width=\textwidth]{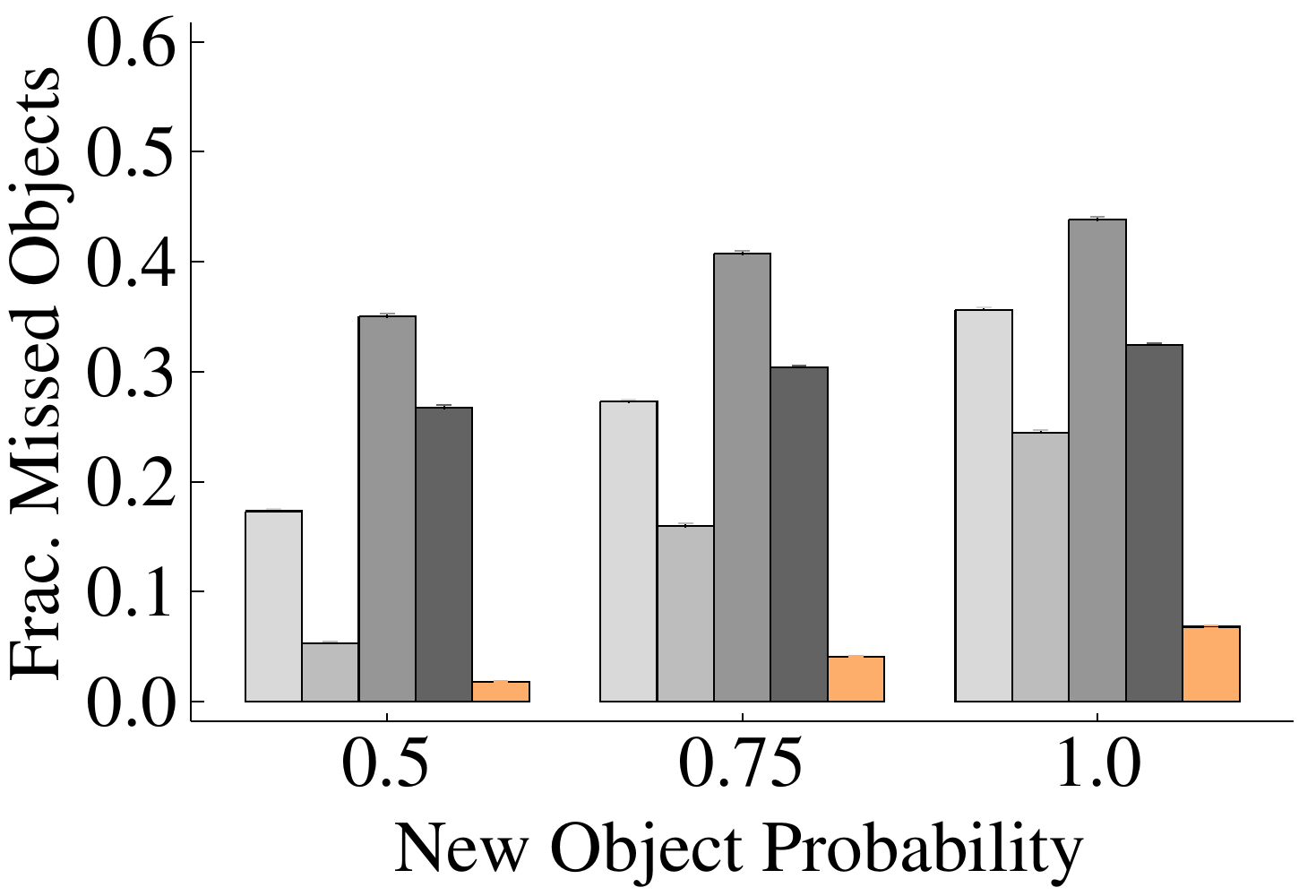}
        \label{fig:belt-perf-box}
    \end{subfigure}
    \caption{Legend: \protect\tikz{\protect\node[fill=mygrey2,draw=black]{};}\; EDD \protect\tikz{\protect\node[fill=mygrey3,draw=black]{};}\; Hungarian 
    \protect\tikz{\protect\node[fill=mygrey4,draw=black]{};}\; MCTS \protect\tikz{\protect\node[fill=mygrey5,draw=black]{};}\; Q-Learning 
    \protect\tikz{\protect\node[fill=myorange,draw=black]{};}\; SCoBA. On the metric of the fraction of unsuccessful tasks, i.e. objects missed, SCoBA consistently outperforms all other baselines. All results are averaged over $100$ trials, with $\horizon = 500$ time-steps per trial.}
    \label{fig:belt-perf}
    \vspace{-10pt}
\end{figure*}

Three robot arms are arranged along a moving conveyor belt, picking objects 
from the belt and placing them in collection bins (\Cref{fig:page1_fig_conv}). We design an abstracted simulation of the
scenario (\Cref{fig:beltviz}), scaled along an X-axis of unit length. The arms have mutually exclusive adjacent
workspaces of $0.3$ units each, from $x = 0.05$ to $x = 0.95$. New tasks arrive as new objects appear at the head of the belt.
Three scenario parameters instantiate the problem and affect the difficulty:
\begin{itemize}
    \item \emph{Grasp Success Probability}: We model uncertainty over task completion due to imperfect grasping with a Bernoulli process where $p_i$ is the probability of a successful pick by the $i$th arm. 
    Accordingly, the cumulative distribution from~\Cref{eq:cumul} is
    \begin{equation}
        \label{eq:cumul-conv}
        \tau_{\agent\task}(t) = 1 - (1 - p_i)^t, \ \ t \in \mathbb{N}.
    \end{equation}
    We expect performance to improve as $p_i$ increases.
    \item \emph{Belt Speed}: The speed of the belt determines the effective time window for each
    arm-object pair, e.g., $\SI{5}{\second}$ in~\Cref{fig:beltviz}. If task execution is successful, each arm has a downtime of $\Delta t = \SI{2}{s}$ to deposit the object in the bin. We expect the performance to degrade as speed increases.
    \item \emph{New Object Probability}: We briefly describe how new objects arrive (more details in the appendix). We reflect the setup in space (about Y-axis) and time, with virtual arms operating in reverse, transferring objects from virtual bins to a virtual belt. Upon crossing the Y-axis, the virtual belt becomes the true belt and virtual objects appear as real ones. The new object probability parameter is the per-timestep Bernoulli probability with which a virtual arm drops its object onto the virtual belt (all virtual arms have the same such probability). The drop location is sampled uniformly within the virtual arm's workspace. As soon as the virtual arm drops an object , it moves to the virtual bin to collect the next one. We expect performance to degrade as new object probability increases.
\end{itemize}
\noindent
\textbf{Competitive Performance against Oracle:}
A standard metric for online algorithms is the competitive performance against an oracle with complete lookahead. The task generation process allows an ablation study for the effect of lookahead alone (decoupled from the effect of uncertainty). If \emph{grasping is perfect}, i.e., $p_i = 1$ for all arms, there exists an oracle strategy that can complete all tasks successfully. The oracle is that which mirrors the generation sequence itself in space and time. We compare SCoBA (without full lookahead) to this oracle by evaluating the proportion of tasks it fails to complete with this generation process. \emph{The smaller this number, the better is SCoBA's competitive performance}.
\begin{table}
\caption{
   Average proportions of objects lost per trial by SCoBA when grasping is perfect. The two varied parameters affect SCoBA's lookahead. The negligible values demonstrate the strong competitive performance of SCoBA relative to the oracle.
    }
    \centering
    \begin{tabular}{@{} lrrr @{}}
        \toprule
                       Belt Speed & \multicolumn{3}{c}{New Object Probability}\\
        \cmidrule{2-4} (units/s)  & $0.5$ & $0.75$ & $1.0$ \\
        \midrule
        $0.04$  & $0.0$     & $0.0$     & $0.0$   \\
        $0.07$  & $\num{2.7e-5}$   & $\num{7.9e-5}$   & $\num{1.3e-4}$ \\
        $0.1$   & $\num{1.7e-4}$   & $\num{3.7e-4}$   & $\num{5.2e-4}$\\
        \bottomrule
    \end{tabular}
    \label{tab:oracle}
    \vspace{1pt}
\end{table}
We set $p_i = 1$ for all arms and jointly vary the other two parameters, belt speed and new object probability. We choose a maximum belt speed of $0.1$ units/s so that each object spends at least $\SI{2}{\second}$ in an arm's workspace. For each trial in a setting, we simulate $\horizon=500$ time-steps (seconds) and evaluate the proportion of objects missed by SCoBA relative to the total number of objects. We compute the average of this proportion-per-trial over $100$ trials (standard error negligible) in~\Cref{tab:oracle}. The low magnitudes demonstrate SCoBA's robustness to insufficient lookahead. With increasing value of either parameter, performance degrades.
\\
\noindent
\textbf{Unsuccessful Task Penalty:}
We vary all three scenario parameters independently and compare the fraction of missed objects for SCoBA versus the other baselines.~\Cref{fig:belt-perf} demonstrates the results.
For each subplot, only one
parameter varies (the x-label), while the other two stay at their default values -- grasp probability $p_i = 0.75$
for all arms, $0.07$ units/s for belt speed, and $0.75$ for new object probability. We average all numbers over $100$ trials (with standard error bars).

\emph{SCoBA considerably outperforms the other baselines across all settings}. Furthermore, its performance degrades or improves
as expected relative to the change in each problem parameter (e.g., more objects missed with increasing new object probability). Among the baselines, the reactive Hungarian method has the best performance, likely because sequential deliberation is not as crucial with non-overlapping workspaces and small downtime (unlike in the next domain). For Q-Learning and MCTS, the performance depends on how finely the conveyor belt is discretized. In Q-Learning, the entire state space also needs to be explicitly enumerated.
We used a discretization of $0.05$ units for Q-Learning and $0.02$ units for MCTS; too much finer would
result in a prohibitively large state space due to the curse of dimensionality~\cite{kochenderfer2015decision}.
\\
\noindent
\textbf{Scalability:}
In this domain, the Coordination Graph is a directed chain (see~\Cref{sec:approach}), so the
computational bottleneck for SCoBA is the policy tree search (multi-agent coordination is trivial). In~\Cref{tab:tree-scalability}
we report average tree search computation times for a single arm with an increasing number of
objects (scattered throughout the arm workspace). Empirically, we observe
that the number of tree nodes is roughly quadratic, and the computation time roughly cubic
in the number of objects, and the wall clock times are quite reasonable.
The appendix has further timing results for the applicable baselines.

\begin{table}
    \centering
    \caption{
    The low computation time values demonstrate that the tree search for an individual arm is quite scalable with respect to the number of objects in the arm's workspace. }
    \begin{tabular}{@{} lcrcr @{}}
        \toprule
        Objects & \phantom{a} & Tree Size & \phantom{a} & Comp. Time\\
        \midrule
        $40$    && $640.9$   && $\SI{9e-4}{\second}$ \\
        $80$    && $2215.3$  && $\SI{0.004}{\second}$ \\
        $120$   && $5102.3$  && $\SI{0.013}{\second}$ \\
        $160$   && $8791.7$  && $\SI{0.029}{\second}$ \\
        $200$   && $13028.4$ && $\SI{0.051}{\second}$ \\
        \bottomrule
    \end{tabular}
    \label{tab:tree-scalability}
    \vspace{1pt}
\end{table}

\begin{figure*}[t]
    \centering
    \begin{subfigure}{0.3\textwidth}
        \centering
        \includegraphics[width=\textwidth]{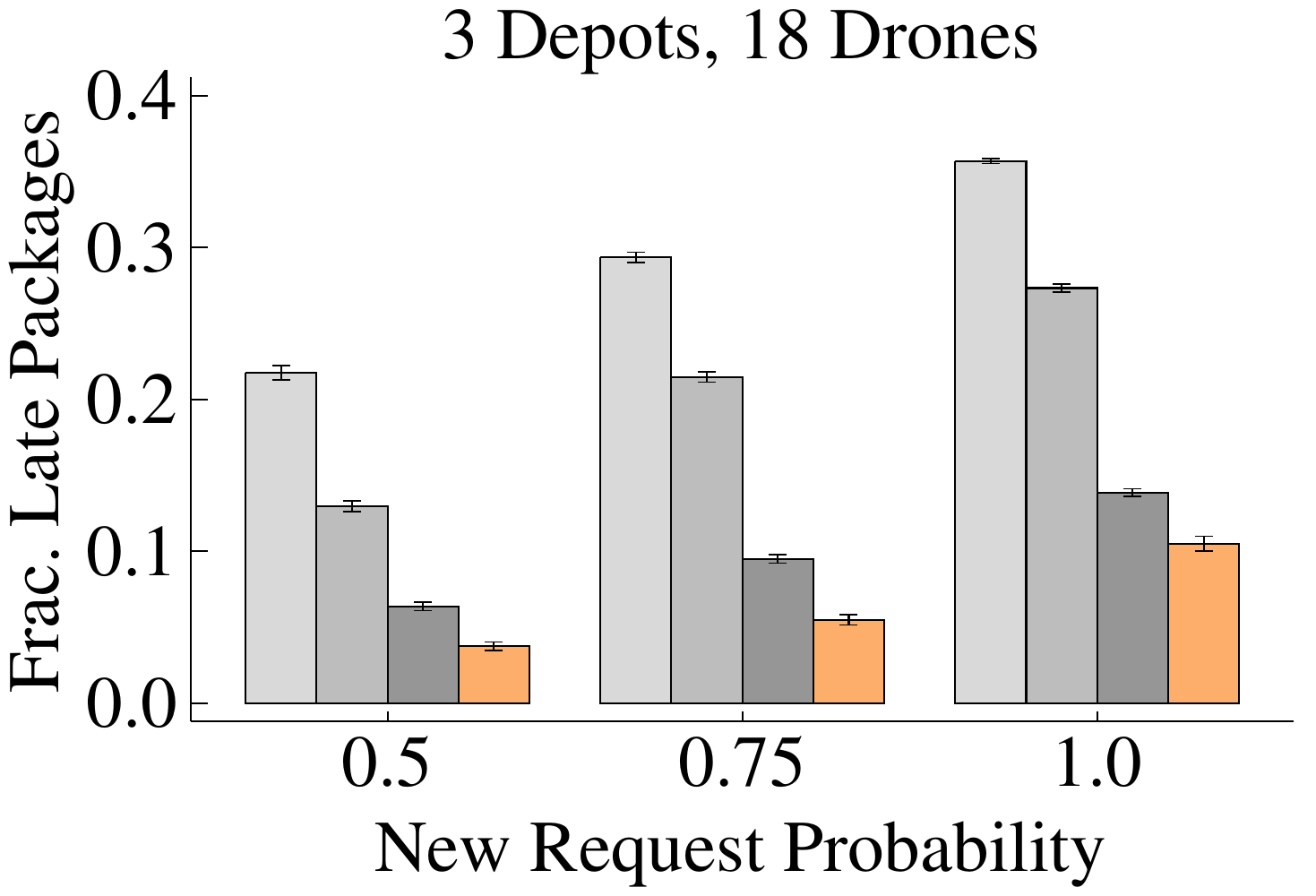}
    \end{subfigure}
    \begin{subfigure}{0.3\textwidth}
        \centering
        \includegraphics[width=\textwidth]{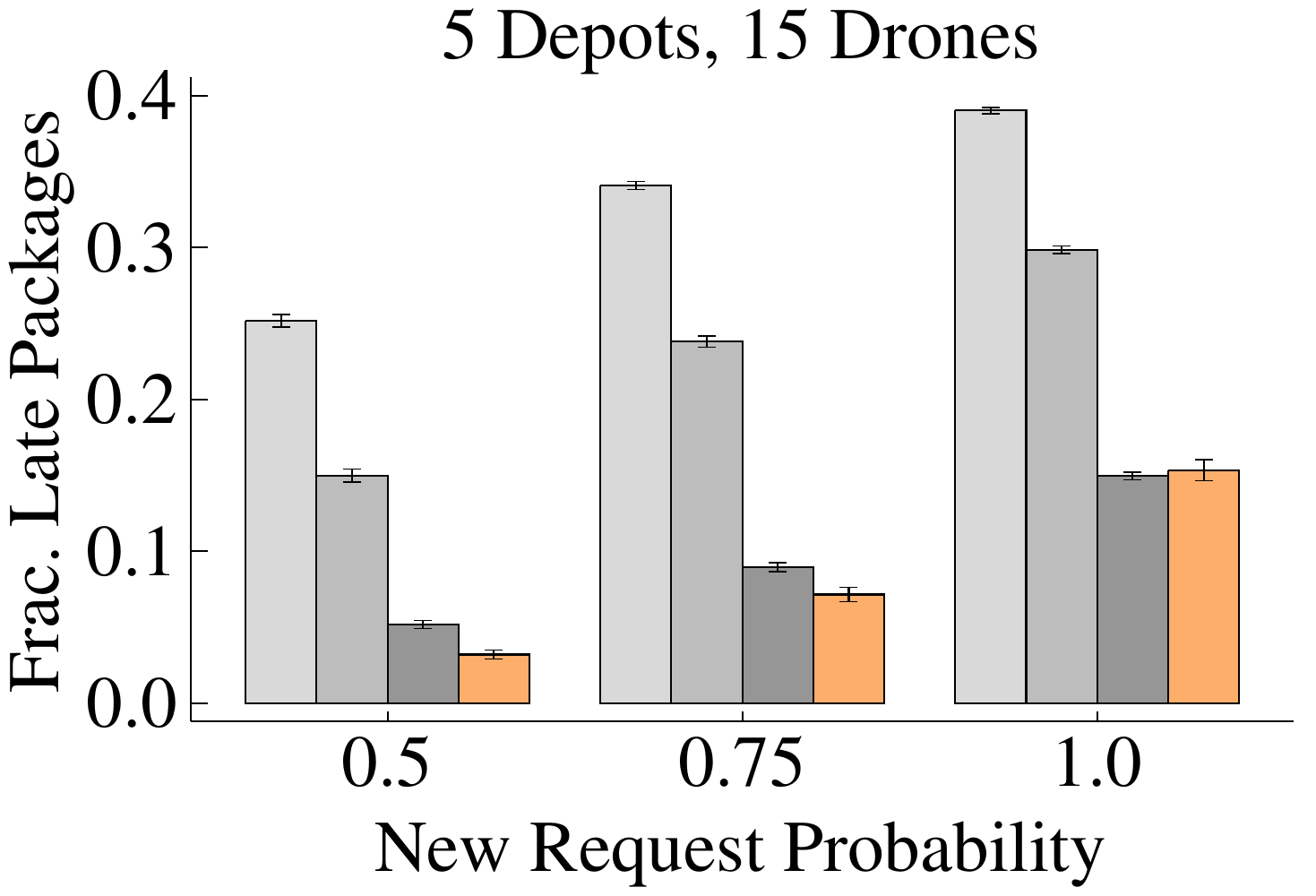}
    \end{subfigure}
    \begin{subfigure}{0.3\textwidth}
        \centering
        \includegraphics[width=\textwidth]{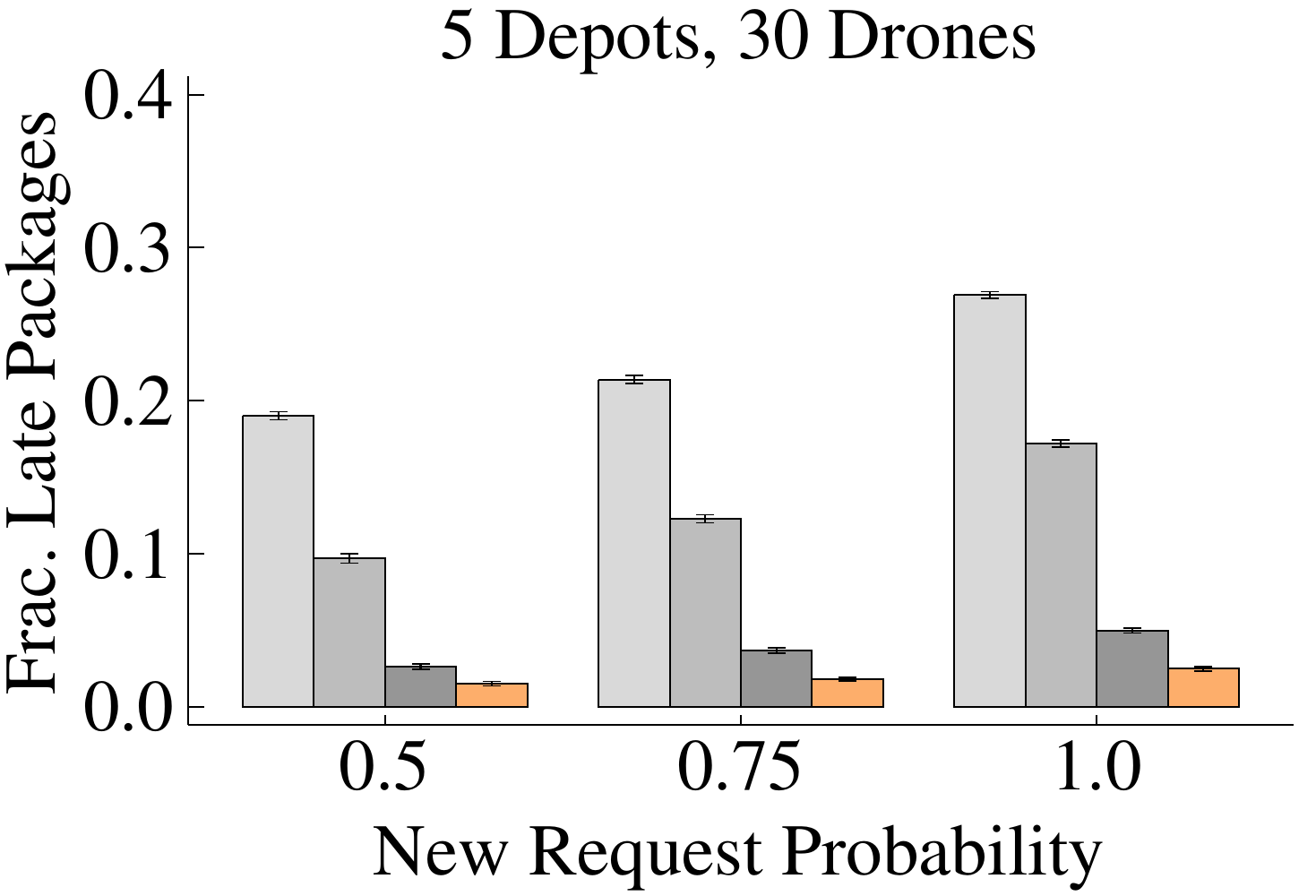}
    \end{subfigure}
    \caption{Legend: \protect\tikz{\protect\node[fill=mygrey2,draw=black]{};}\; EDD \protect\tikz{\protect\node[fill=mygrey3,draw=black]{};}\; Hungarian 
    \protect\tikz{\protect\node[fill=mygrey4,draw=black]{};}\; MCTS \protect\tikz{\protect\node[fill=myorange,draw=black]{};}\; SCoBA.
    For the drone delivery domain, on the primary metric of the fraction of late package deliveries, SCoBA outperforms the baselines on all but one setting.
    Results are averaged over $100$ trials each of $\horizon = 100$ time-steps.
    }
    \label{fig:routing_perf_plots}
    \vspace{-10pt}
\end{figure*}

\subsection{Drone Delivery: Experiments and Results}
\label{sec:eval-deliv}

In the second domain, we dispatch drones to deliver packages around a city-scale area, subject
to delivery time window constraints. Our setup is based on our recent work for multi-drone delivery over
ground transit~\cite{choudhury2019efficient}. We use the location-to-location travel time estimates from its North San Francisco scenario, which simulates deliveries over an area of \SI{150}{\kilo \metre \squared} (see~\Cref{fig:page1_fig_drone}).  We handpick locations for up to $5$ depots scattered around the city to ensure good coverage. Drones have a maximum flight range of \SI{10}{\kilo \metre}, which restricts the set of possible package deliveries for each drone.
Two scenario parameters affect the performance here:
\begin{itemize}
    \item \emph{Drones and Depots}: The number of depots and the ratio of drones to depots both impact the ability of the system to dispatch agents to a given delivery location in time. We distribute drones equally across depots. With better coverage,
    we expect performance to improve.
    \item \emph{New Request Probability}: A new delivery request arrives per minute with some probability. Each delivery location is sampled uniformly within a bounding box. We start with a number of packages roughly $1.5$ times the number of drones in the scenario. With higher probability, we expect performance to degrade. Each request has a window duration sampled uniformly between $15$ and $30$ minutes. 
\end{itemize}

Our reference framework gives us deterministic drone travel-time estimates between a depot $d$ and 
package location $p$, say $TT(d,p)$. We model travel time uncertainty with a finite-support
Epanechnikov distribution around $TT(d,p)$, i.e.,
\begin{equation}
    \label{eq:cumul-deliv}
    \tau_{\agent,\task}(t) \sim \mathrm{Epan}(\mu = TT(d,p), r = TT(d,p)/3.0).
\end{equation}
The true travel time is drawn from~\Cref{eq:cumul-deliv}. This choice of a synthetic distribution is arbitrary but reasonable because a high mean travel time is likely to have higher variance, due to more opportunities for delays or speedups.

\subsubsection{Unsuccessful Task Penalty}

We vary the scenario parameters and compare
the fraction of late package deliveries for SCoBA versus the other baselines in~\Cref{fig:routing_perf_plots}. We choose three sets of depot-and-drone numbers with complementary coverage properties, e.g., $(3,18)$ has fewer depots and a higher drone-depot ratio while $(5,15)$ has more depots but a smaller ratio. We vary the new request probability, simulate $\horizon = 100$ time-steps (minutes) per trial, and average results over $100$ trials. We omit Q-Learning because the enumeration of the multi-agent MDP state space is unacceptably large for any useful time-axis discretization.
SCoBA is generally the best across all settings except in one ($5$ depots, $15$ drones, probability $1.0$) where MCTS is slightly better. 
The vast improvement in relative performance of the MCTS baseline~\cite{DBLP:conf/aaai/KartalNGG16} is not surprising. It is tailored for vehicle dispatch problems, 
with search heuristics that exploit the domain structure, e.g., agents have longer downtime to return to their depots; the per-agent action space is a subset of
valid tasks rather than a discretization of a conveyor belt into slots.
For the coverage parameter, having more drones per depot appears to be more influential than having more depots, e.g., the errors for $(5,15)$ are higher than the corresponding ones for $(3,18)$.

\subsubsection{Scalability}

In this domain, high-level conflicts may occur, and SCoBA will invoke its multi-agent coordination layer (over and above policy tree search) while computing a valid multi-agent allocation. Therefore, in~\Cref{tab:routing-scale} we report the mean and standard error for computation times for the full SCoBA algorithm (over $50$ different trials for each setting). We vary the number of drones and depots and the number of currently available tasks, i.e., the current package delivery requests. The absolute wall clock values are quite reasonable given that the time-scale of operation of the system in the real world is minutes and hours. Some scenarios have disproportionately high mean and variance because of more high-level conflicts, a known behavioral property of Conflict-Based Search algorithms~\cite{sharon2012conflict}. The appendix has timing results for other baselines.

\begin{table}
\caption{The mean and standard error (over $50$ trials in each setting) for SCoBA computation times on the multi-drone delivery domain. All times are in seconds.}
    \centering
    \begin{tabular}{@{} lrrr @{}}
        \toprule
        & \multicolumn{3}{c}{Number of Requests}\\
        \cmidrule{2-4} (Depots,Drones) & $20$ & $50$ & $100$ \\
        \midrule
        $(3,18)$  & $(0.02,0.003)$   & $(1.48,0.16)$    & $(2.55, 0.23)$   \\
        $(5,15)$  & $(0.06,0.008)$   & $(2.13,0.2)$    & $(5.42,0.5)$ \\
        $(5,30)$  & $(0.17,0.003)$   & $(1.76,0.12)$    & $(7.08,0.47)$\\
        \bottomrule
    \end{tabular}
    \label{tab:routing-scale}
    \vspace{3pt}
\end{table}

\section{Conclusion}
\label{sec:conclusion}

We presented SCoBA, a hierarchical approach for dynamic multi-robot task allocation under uncertainty and temporal constraints.
In theory, SCoBA is optimal in expectation and complete under mild technical assumptions. In practice, over two distinct and realistic
domains, it has strong competitive performance against an oracle, consistently outperforms a number of baselines, and is scalable in terms of computation time to both agents and tasks.
\\
\noindent
\textbf{Limitations and Future Work:}
We assume a known uncertainty model, which is typical for multi-robot task allocation research. However, since SCoBA is based on policy tree search, we could use it in-the-loop with model-based RL in case the uncertainty model needs to be estimated online. SCoBA's computation time is sensitive to the number of high-level conflicts; future work could incorporate
efficiency improvements to Conflict-Based Search such as bounded sub-optimal variants~\cite{DBLP:conf/ecai/BarerSSF14} and improved conflict resolution~\cite{DBLP:conf/ijcai/BoyarskiFSSTBS15}. Finally, we focus on high-level allocation here, but we could 
integrate SCoBA in a full pipeline for robotics applications.

\section*{Acknowledgments}
This work was supported by the Ford Motor Company, NSF grant number 1241349 and NSF grant number 138329.

\printbibliography

\newpage
\clearpage
\appendix

\section*{Optimality and Completeness Proofs}

\noindent
\textbf{Proposition 1.}
\emph{
If (i) no new tasks are added online, (ii) the tree search is executed to the full horizon and (iii) task completion is determined at the end of the time window, then SCoBA is optimal in expectation, i.e. SCoBA minimizes in expectation the number of incomplete tasks at the end of the time horizon.
}
\begin{proof}

Conflict-Based Search has been proved to yield optimal multi-agent solutions (paths) if two conditions hold:
(a) the low-level routine yields optimal solutions for individual agents and 
(b) the overall multi-agent  objective is the sum-of-costs of the individual agent solutions. 
SCoBA uses the same high-level multi-agent conflict resolution logic as Conflict-Based Search, and will inherit this optimality property if it satisfies the two sufficient conditions.

We first show (a) is true for SCoBA.
The low-level single-agent routine uses dynamic programming with forward tree search
to obtain a policy tree. Such a method, by construction, computes
a policy from the initial state that is optimal under expectation for a discrete, finite-horizon Markov Decision Process (MDP) \emph{if the tree search is exhaustively conducted up to the full horizon}. The expectation is 
over the uncertainty of the outcomes of actions.
Assumption (i) ensures that all information of future tasks is known at the initial state of the agent and Assumption (ii)
ensures the exhaustive tree search. 

Recall the simplifying approximation we made along the temporal dimension, by treating time-windows
as discrete events. With Assumption (iii), the temporal approximation now becomes exact. i.e., by attempting each task at the
first possible time-step and continuing until the end of the window, each task attempt has a single probability
mass function over the two possible outcomes, success or fail.
Therefore, the single agent problem can be framed as a discrete finite-horizon MDP and
the low-level policy tree search routine is optimal in expectation for an individual agent. 
That is, for each agent $n$, the policy $\pi^{*}(n)$ obtained from $\textsc{PlanTree}$ satisfies
\begin{equation}
    \label{eq:condtree-opt}
   \pi^{*}(n) = \underset{\pi(\agent) \in \Pi(\agent)}{\mathrm{argmin}} \ \mathbb{E} \big[\sum_{\task \in \pi(\agent)}  \mathds{1}[k] \cdot J(\task) \mid \pi(\agent) \big] 
\end{equation}
subject to the constraints in~\Cref{eq:opt-prob} of the main text, where $\Pi(n)$
is the set of all possible policy trees for agent $n$. With some abuse of notation, we use $k \in \pi(\agent)$ to denote all the tasks allocated to agent $\agent$.
\emph{Thus, SCoBA satisfies  condition (a) from above}.

Now let us show (b) holds for SCoBA.
The true overall objective in SCoBA is the expected cumulative penalty of unsuccessful tasks 
due to the computed multi-agent allocation policy $\pi$. For convenience, denote this objective as $J(\pi)$. Then, once
again from~\Cref{eq:opt-prob} in the main text, we have
\begin{equation}
\label{eq:trueobj}
J(\pi) = \mathbb{E} \big[\sum_{\task \in [\tasks]} \mathds{1}[k] \cdot J(\task) \mid \pi \big].
\end{equation}
By the linearity of expectation, we rewrite~\Cref{eq:trueobj} as
\begin{equation}
\label{eq:sumofexp}
J(\pi) =  \sum_{\task \in [\tasks]}  \mathbb{E} \big[ \mathds{1}[k] \cdot J(\task) \mid \pi \big].
\end{equation}
By construction, SCoBA resolves conflicts between single-agent allocation policies to ensure that no two agents are allocated to the same task. Therefore, we can split the summation over all tasks by the tasks allocated to agents and rewrite~\Cref{eq:sumofexp} as
\begin{equation}
\label{eq:sumofexpperagent}
J(\pi) = \sum_{\agent} \sum_{\task \in \pi(n)} \mathbb{E} \big[ \mathds{1}[k] \cdot J(\task) \mid \pi(n) \big].
\end{equation}
where, once again, we use $k \in \pi(\agent)$ to denote all the tasks uniquely
allocated to agent $\agent$.

Now, consider the objective function for the single-agent policy in~\Cref{eq:condtree-opt}. Recall from~\Cref{sec:approach-tree} of the main text that  \emph{the minimizing value of the single-agent objective is} $V(\mathrm{root}_n)$
where $\mathrm{root}_{\agent}$ is the root of the single-agent optimal plan tree $\pi^{*}(\agent)$, i.e.
\begin{equation}
    \label{eq:condtree-optval}
   V(\mathrm{root}_n) = \underset{\pi(\agent) \in \Pi(\agent)}{\mathrm{min}} \ \mathbb{E} \big[\sum_{\task \in \pi(\agent)} \mathds{1}[k] \cdot J(\task) \mid \pi(n) \big].
\end{equation}
By using linearity of expectation again, we can write~\Cref{eq:condtree-optval} as
\begin{equation}
    \label{eq:condtree-optval-exp}
   V(\mathrm{root}_n) = \underset{\pi(\agent) \in \Pi(\agent)}{\mathrm{min}} \sum_{\task \in \pi(\agent)} \mathbb{E} \big[\mathds{1}[k] \cdot J(\task) \mid \pi(\agent) \big].
\end{equation}

Finally, using~\Crefrange{eq:sumofexpperagent}{eq:condtree-optval-exp} and that no two agents are allocated to the same task, we have

\begin{equation}
\begin{aligned}
     \min_{\pi \in \Pi} J(\pi) &= \min_{\pi \in \Pi} \sum_{\agent} \sum_{\task \in \pi(n)} \mathbb{E} \big[\mathds{1}[k] \cdot J(\task) \mid \pi(n) \big] \\
     &= \sum_{\agent} \min_{\pi(\agent) \in \Pi(\agents)} \sum_{\task \in \pi(n)} \mathbb{E} \big[\mathds{1}[k] \cdot J(\task) \mid \pi(n) \big] \\
     &= \sum_{\agent} V(\mathrm{root}_n)
\end{aligned}
\end{equation}
which is precisely the sum of costs of the individual agent solutions. Thus, \emph{SCoBA
satisfies condition (b) from above}.
  
Therefore, SCoBA is optimal in expectation under the given assumptions.
\end{proof}

\noindent
\textbf{Proposition 2.}
\emph{
Under the same assumptions as Proposition 1, SCoBA is complete. If a conflict-free allocation exists,
SCoBA will return it.
}

\begin{figure*}[th]
    \centering
    \includegraphics[width=0.9\textwidth]{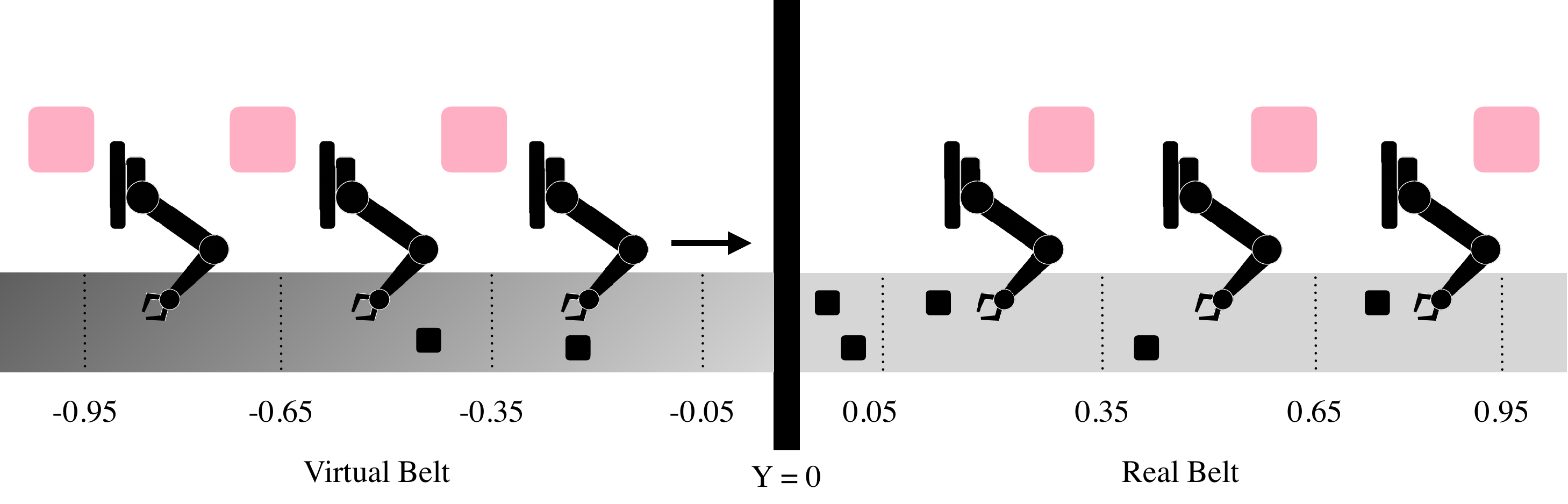}
    \caption{The task generation process for the conveyor belt domain is defined by reflecting the real setup in space and time to obtain a virtual setup.}
    \label{fig:beltvizmirror}
\end{figure*}

\begin{proof}
As with the proof for optimality, our proof for completeness follows that for the original Conflict-Based Search algorithm.
In the case of Conflict-Based Search, completeness was shown by establishing that the high-level constraint
tree has a finite number of nodes, i.e., an upper bound on how many nodes can be generated, if a valid multi-agent
path does exist. Because Conflict-Based Search generates at least one new constraint per new high-level node, and executes best-first search with monotonically non-decreasing cost on the constraint
tree, it is guaranteed to find a valid solution in finite time if one exists.

Let us now apply the same reasoning to SCoBA.
First, every new high-level node $A$ in SCoBA's constraint tree must have at least one additional constraint as compared to
its predecessor $A'$, derived from a conflict in the solution that $A'$ represents. 
Second, the total number of possible constraints
is finite. Specifically, the maximum possible number of constraints is the number of ways $\tasks$ tasks can be distributed across
$\agents$ agents (obtained from standard combinatorics results)
multiplied by the time horizon, i.e. $\frac{(\tasks + \agents - 1)!}{\tasks ! (\agents - 1)!} \cdot T$.
Therefore, the finite number of possible constraints and the addition of at least one new constraint per new node implies that there is a finite number of nodes in the constraint tree. 
The high level routine in SCoBA uses systematic best-first search
over the constraint tree, whose expanded nodes have monotonically non-decreasing cost (by construction).
Therefore, a conflict-free allocation, if it exists, must be found after expanding a finite
number of SCoBA constraint tree nodes. Thus, SCoBA is a complete algorithm.
\end{proof}

\section*{Additional Evaluation Details}

\subsection*{Conveyor Belt Task Generation Process}

\noindent
We discuss in more detail the conveyor belt task generation process mentioned in~\Cref{sec:eval-convbelt} of the main text.~\Cref{fig:beltvizmirror}
provides a supporting illustration for the same. We reflect the belt setup in both space and time and create a virtual assembly line (on the left side of the
figure) where arms pick up objects from bins and place them on the virtual belt. When the virtual belt crosses over, the virtual
objects appear as new real objects.

Recall that the above task generation process, by construction, implies the existence of at least one allocation strategy that is perfect, i.e., that 
successfully executes all tasks when there is no uncertainty. One perfect allocation strategy is that which reflects the virtual setup that
generates the tasks. By this strategy, for instance, the first arm, i.e., the arm closest to the belt origin, is allocated to pick up those real objects
whose corresponding virtual forms were placed on the belt by the arm's reflection. The attempt location is the reflection (on the real belt) of the point in the
virtual arm's workspace where it drops the virtual object.

In practice, of course, we would require complete access to the virtual generator in order to obtain the perfect allocation strategy (assuming no uncertainty) for
a given problem instance. However, note that evaluating the competitive performance of an online algorithm with respect to an oracle (as we do in the main text)
only requires us to know the performance of an oracle, which is an upper bound on the performance of any other method.
The existence of a perfect allocation strategy implies that an oracle with access to all future information would have full success rate for any task sequence
generated by our process. We can thus effectively compute the competitive performance of SCoBA (or any other approach) simply by evaluating the fraction of 
unsuccessful tasks (missed boxes) of that approach.

\subsection*{Baseline Implementation Details}

\noindent
Like SCoBA, all baselines were implemented in the Julia language~\cite{Julia-2017}. For both MCTS and Q-Learning, we used the implementations in the POMDPs.jl framework~\cite{egorov2017pomdps} for
modeling and solving Markov Decision Processes. For the rollout policy of MCTS, we used the EDD heuristic itself, which was better than random rollout.
For Q-Learning, we used $\epsilon$-greedy exploration with $\epsilon$ scaled linearly with a decay rate of $0.9995$. The learning rate was $0.01$ and
the policy was trained for $100000$ steps.

\subsection*{Baseline Computation Time Results}

\begin{table}
    \centering
    \caption{Timing comparisons -- Conveyer Belt.}
    \begin{tabular}{@{} lcrcr @{}}
        \toprule
        Objects & \phantom{a} & SCoBA & \phantom{a} & Hungarian\\
        \midrule
        $40$    && $\SI{9e-4}{\second}$ && $\SI{1.7e-5}{\second}$ \\
        $80$    && $\SI{0.004}{\second}$ && $\SI{3.1e-5}{\second}$ \\
        $120$   && $\SI{0.013}{\second}$ && $\SI{4.3e-5}{\second}$ \\
        $160$   && $\SI{0.029}{\second}$ && $\SI{5.7e-5}{\second}$ \\
        $200$   && $\SI{0.052}{\second}$ && $\SI{8.3e-5}{\second}$ \\
        \bottomrule
    \end{tabular}
    \label{tab:tree-baseline-scalability}
\end{table}

\begin{table*}[th]
\caption{Timing comparisons -- Drone Delivery. All times are in seconds.}
    \centering
    \begin{tabular}{@{} lcrrrcrrrcrrr @{}}
        \toprule
        && \multicolumn{3}{c}{$20$ Requests} && \multicolumn{3}{c}{$50$ Requests} && \multicolumn{3}{c}{$100$ Requests}\\
        \cmidrule{3-5} \cmidrule{7-9} \cmidrule{11-13} (Depots,Drones) && SCoBA & Hungarian & MCTS && SCoBA & Hungarian & MCTS && SCoBA & Hungarian & MCTS \\
        \midrule
         $(3,18)$  && $0.02$ & $\num{8.4e-5}$& $0.007$ && $1.48$ & $\num{1e-4}$& $0.008$ && $2.55$ & $\num{2e-4}$& $0.009$ \\
         $(5,15)$  && $0.06$ & $\num{8.3e-5}$& $0.006$ && $2.13$ & $\num{1e-4}$& $0.007$ && $5.42$ & $\num{2e-4}$& $0.009$ \\
         $(5,30)$  && $0.17$ & $\num{2e-4}$& $0.01$ && $1.76$ & $\num{2e-4}$& $0.013$ && $7.08$ & $\num{3e-4}$& $0.016$ \\
        \bottomrule
    \end{tabular}
    \label{tab:routing-scale-baselines}
\end{table*}
\noindent
As we had flagged in the main text, we now report the computation times of some baseline approaches in our two simulation domains. The comparison is not apples-to-apples, because the baselines have different input interfaces that impose different restrictions on the full problem size, depending on the domain (for instance, the MDP approaches in the conveyor belt domain are unaffected by the number of tasks due to the action space being the belt slots obtained by discretizing it). Furthermore, computation time for us is not an optimizing metric but rather a satisficing one; our objective for SCoBA's computation time is to be reasonable for the requirements of the respective domains (which they are, as we discuss in the main text).
\\
\\
\noindent
\textbf{Conveyor Belt:} In~\Cref{tab:tree-baseline-scalability} we compare the computation time of the Hungarian baselines
with that of SCoBA's policy tree search, with varying objects. The times in the SCoBA column are 
copied over from~\Cref{tab:tree-scalability}.
Clearly, Hungarian approach is much faster than SCoBA, but unlike SCoBA it does not plan sequentially and only matches an arm to objects within the arm workspace (so the effective number of objects considered per arm is about a third of the total number, for each of the three arms).
For MCTS, the action computation time is independent of the number of the objects because the action
space is the discretization of the conveyor belt into $15$ slots per agent.
With $100$ MCTS iterations and a search depth of $20$, the average action computation time is $\SI{0.1}{\second}$.

The computation times for EDD and Q-Learning are not informative; EDD is a simple heuristic that does not plan jointly for all agents, and in Q-learning the tabular policy is precomputed offline and simply looked up online (and like MCTS the action space is a discretization of belt slots and does not depend on the number of objects).
\\
\\
\noindent
\textbf{Drone Delivery:}
In~\Cref{tab:routing-scale-baselines}, we compare mean computation times of the Hungarian and MCTS baselines to those of the full SCoBA algorithm, with varying depots, drones, and requests considered. The values in the SCoBA sub-columns are copied over from the corresponding entries in~\Cref{tab:routing-scale} (the first quantity of the mean and standard error tuple). As expected, Hungarian is orders of magnitude faster than SCoBA. For the MCTS baseline, the action space is now directly proportional to the number of tasks, i.e., requests considered, and the computation times do vary accordingly. The absolute numbers are much lower than SCoBA primarily due to the MCTS baseline circumventing the complexity of multi-agent coordination by imposing an arbitrary ordering on the assignment of agents.

\end{document}